\documentclass[journal]{IEEEtran}

\usepackage{enumerate}
\usepackage[table]{xcolor}
\usepackage[noadjust]{cite}
\usepackage[cmex10]{amsmath}
\usepackage{subfigure}
\usepackage{graphicx, mdwtab}
\graphicspath{{Figures/}}
\usepackage{placeins}
\usepackage{afterpage}
\usepackage{multirow}
\usepackage[font=small]{caption}%,labelsep=qquad
\DeclareCaptionLabelSeparator{qquad}{.~~}
\begin{document}
\title{Hole Filling with Multiple Reference Views in DIBR View Synthesis}

\author{Shuai~Li, Ce~Zhu,~\IEEEmembership{Fellow,~IEEE}, Ming-Ting~Sun,~\IEEEmembership{Fellow,~IEEE}%\textsuperscript{*}
         % <-this % stops a space
%\thanks{\textsuperscript{*}Correspondence: eczhu@uestc.edu.cn.}
\thanks{Manuscript received June 25, 2017. This work was supported in part by the National Natural Science Foundation of China under Grant 61571102, in part by the Fundamental Research Funds for the Central Universities under Grant ZYGX2014Z003, and in part by the National High Technology Research and Development Program of China under Grant 2015AA015903. \textit{Corresponding author: Ce Zhu}}%  %This work was supported in part by the National High Tech. R\&D Program of China (No. 2015AA015903) and Fundamental Research Funds for the Central Universities (No. ZYGX2014Z003).}% <-this % stops a space
\thanks{S. Li and C. Zhu are with School of Electronic Engineering, University of Electronic Science and Technology of China, Chengdu, China. Email: shuailichn@gmail.com; eczhu@uestc.edu.cn.}% <-this % stops a space    %611731
\thanks{Ming-Ting Sun is with the University of Washington, Seattle, USA. E-mail: mts@uw.edu.}
}

\maketitle

\begin{abstract}%
Depth-image-based rendering (DIBR) oriented view synthesis has been widely employed in the current depth-based 3D video systems by synthesizing a virtual view from an arbitrary viewpoint. However, holes may appear in the synthesized view due to disocclusion, thus significantly degrading the quality. Consequently, efforts have been made on developing effective and efficient hole filling algorithms. Current hole filling techniques generally extrapolate/interpolate the hole regions with the neighboring information based on an assumption that the texture pattern in the holes is similar to that of the neighboring background information. However, in many scenarios especially of complex texture, the assumption may not hold. In other words, hole filling techniques can only provide an estimation for a hole which may not be good enough or may even be erroneous considering a wide variety of complex scene of images. In this paper, we first examine the view interpolation with multiple reference views, demonstrating that the problem of emerging holes in a target virtual view can be greatly alleviated by making good use of other neighboring complementary views in addition to its two (commonly used) most neighboring primary views. The effects of using multiple views for view extrapolation in reducing holes are also investigated in this paper. In view of the 3D Video and ongoing free-viewpoint TV standardization, we propose a new view synthesis framework which employs multiple views to synthesize output virtual views. Furthermore, a scheme of selective warping of complementary views is developed by efficiently locating a small number of useful pixels in the complementary views for hole reduction, to avoid a full warping of additional complementary views thus lowering greatly the warping complexity. Experimental results show that the hole size based on two primary reference views may be reduced by up to about 70\% with the help of two complementary reference views in the case of view interpolation, while the hole size based on one primary reference view may be reduced by about 27\% with the help of one more complementary reference view in view extrapolation. Moreover, it is shown that by using one more pair of views in view interpolation and one more view in view extrapolation, 10\% hole pixels may be reduced additionally.
\end{abstract}


% \fi
\IEEEpeerreviewmaketitle

\section{Introduction}

\IEEEPARstart{F}{ree}-viewpoint video \cite{SP_FTV, Proceedings_FTV} is an advanced visual media type and has been widely recognized as the next generation video application, as it can provide users immersive 3D feelings while watching video. Great efforts have been put into investigating the realization of high-quality free-viewpoint video and the Moving Picture Experts Group (MPEG) has been conducting Free-viewpoint TV (FTV) standardization \cite{MPEG_FTV} since 2001. In 2001, FTV was proposed to MPEG and the corresponding 3D Audio Visual (3DAV) activity started. The first phase of FTV, which is Multi-view Video Coding (MVC), was initiated in 2004 and completed in 2009. The second phase of FTV, which is known as 3D Video (3DV), started in 2007 and just completed recently. In the recent MPEG meetings since July 2014, FTV has started a new round of exploration experiments \cite{MPEG_EE} for the third phase of FTV.

Each phase has its own target with applications in accordance with the technologies of its time. Among them, MVC has been adopted by Blue-ray 3D, which involves only (texture/color) video of multiple views and exploits the correlation among different views to further enhance the coding efficiency. It can provide the user a few predefined viewing angles for watching. The current 3DV involves both texture videos and the corresponding depth videos of multiple views, e.g. 3 views of texture and depth videos in the recommended configuration. These views are sent in the encoder side while a larger number of views can be generated at the receiver side based on these views by employing view synthesis. It generally targets the multiview displays with views less than about 30. The newly started FTV exploration experiment aims at two specific applications, super multiview video and free viewpoint navigation, which requires the system to be able to synthesize dense views and specified views. Both the current 3DV and the newly started FTV need to provide virtual views using the received multiple views (decoded on the user side), which makes the view synthesis a key component in the processing chain of the 3D video system.

Depth-image-based rendering (DIBR) oriented view synthesis \cite{Zhu_book,DIBR} is one of the representative view synthesis methods \cite{lightfield2006, Debevec1996IBR, Lipski2014IBR, 2011Kinectfusion} due to its capability in rendering virtual views at arbitrary viewpoints. An inherent problem in DIBR view synthesis is that the regions occluded by the foreground objects in the reference view may become visible in the synthesized view. These exposed areas known as holes in the virtual view will greatly degrade the quality of the synthesized image if not being dealt with properly. In a typical view interpolation by utilizing two nearst neighboring views (noted as primary views throughout this paper) to synthesize an in-between virtual view, the disocclusion problem can be alleviated to some extent as uncovered regions in one view may be visible in the other view and thus some of the holes may be filled by merging the two different synthesized views. However, some holes may still remain in the merged view.

There are currently some works on hole filling \cite{Barnes2009Inpaint, Kumar2005Inpaint, Telea2004Inpaint,SPIC_FTV, schmeing2015faithful, ndjiki2011depth, Jour_Hole_tempor,  Myconf_IIHMSP, MyJour_TBC} available in the literature. Most of them employ image inpainting techniques \cite{Barnes2009Inpaint, Kumar2005Inpaint, Telea2004Inpaint,SPIC_FTV,schmeing2015faithful} as they have been widely used to fill cracks or selected regions of an image. Considering that holes generated in view synthesis are exposed regions occluded by foreground objects, filling holes with background information is more plausible. In \cite{ndjiki2011depth}, a depth based inpainting method was proposed for hole filling where the non-hole regions of smaller depth values neighboring to the holes are regarded as background and used for filling up the holes. However, there are cases where no background information is available near a hole region, and such a method may fail. In \cite{Myconf_IIHMSP,MyJour_TBC}, the view interpolation cases are examined while the distribution of the background around a hole is thoroughly studied. Accordingly a hole filling method was proposed based on occluded information which belongs to background. There are also methods \cite{Jour_Hole_tempor} exploiting temporal correlation information to fill up holes, which may not be applicable for image view synthesis discussed in this paper. Though relatively good performance are shown in these works under some circumstances, they may not perform well in the cases when the holes are large and/or the texture of the hole regions is too complex to be predicted with the neighboring information. The hole filling techniques only provide an estimation for the hole regions based on the assumption that the pixels of the hole regions are similar to its neighboring non-hole regions in an image or non-hole regions in the neighboring temporal frames. Accordingly, quality of a filled region is constrained by how well the assumption may hold.

On the other hand, in the ongoing 3DV system or the newly started FTV system, generally at least three views of both texture and associated depth videos are coded and sent to the decoder side, which may provide more information for view synthesis with all the texture and depth videos. In our previous work \cite{Myconf_BMSB}, we proposed to use multiple reference views for hole reduction which can efficiently reduce the holes. In this paper, it is further developed by proposing a new framework which takes all the available views into consideration in synthesizing virtual views. Moreover, hole reduction with complementary views in view extrapolation is discussed as well as in view interpolation. It is shown and verified in experiments that using all the available views can assist reducing holes while synthesizing virtual views. The main contributions of our paper are summarized as follows.

(1) We perform a thorough examination of the hole generation process with multiple reference views in the DIBR view interpolation and view extrapolation, respectively.

(2) We obtain analytical results that the complementary views are useful for hole reduction under certain circumstances and the reduction effect can be calculated quantitatively. Accordingly, a new view synthesis framework exploiting all possible views available is developed.

(3) We develop a selective warping scheme, which only selects the most relevant pixels (instead of full warping) of the complementary views, to lower the rendering complexity.

The organization of the paper is as follows. Section \ref{MPtheory} thoroughly examines the mechanism of hole generation with multiple reference views in view interpolation and view extrapolation, respectively. Section \ref{selectivewarp} presents the proposed selective warping scheme which selects only relevant pixels in the complementary views in the warping. Experimental results are shown in Section \ref{experiment}, and conclusion remarks are drawn in Section \ref{conclusion}.

\begin{figure}[tbp]
\centering
\includegraphics[width=0.8\hsize]{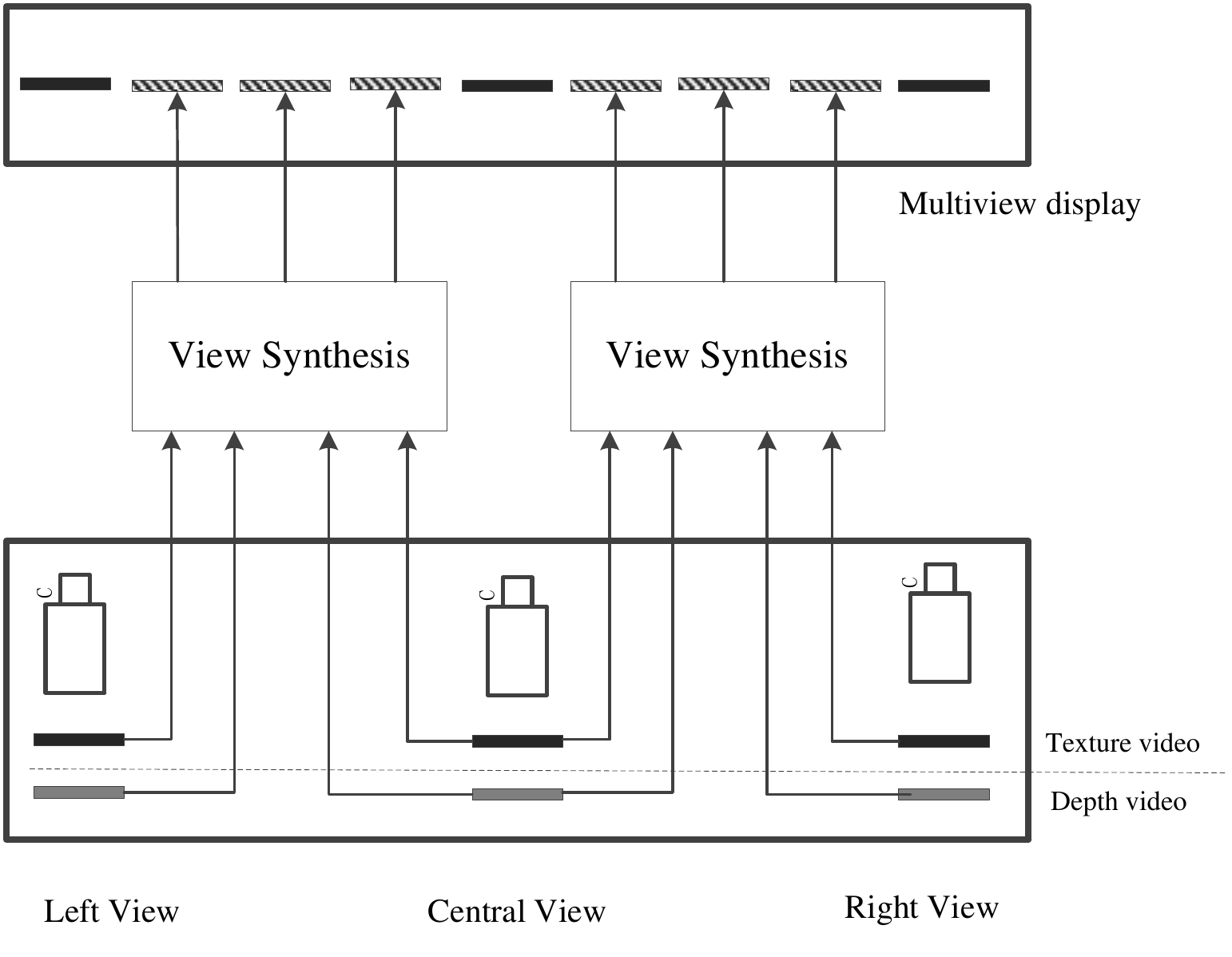}
\caption{An example configuration for 3D video with texture and depth videos of three views.}
\label{fig_3dv}
\end{figure}

\begin{figure*}[tbp]
\centering
\includegraphics[width=1\hsize]{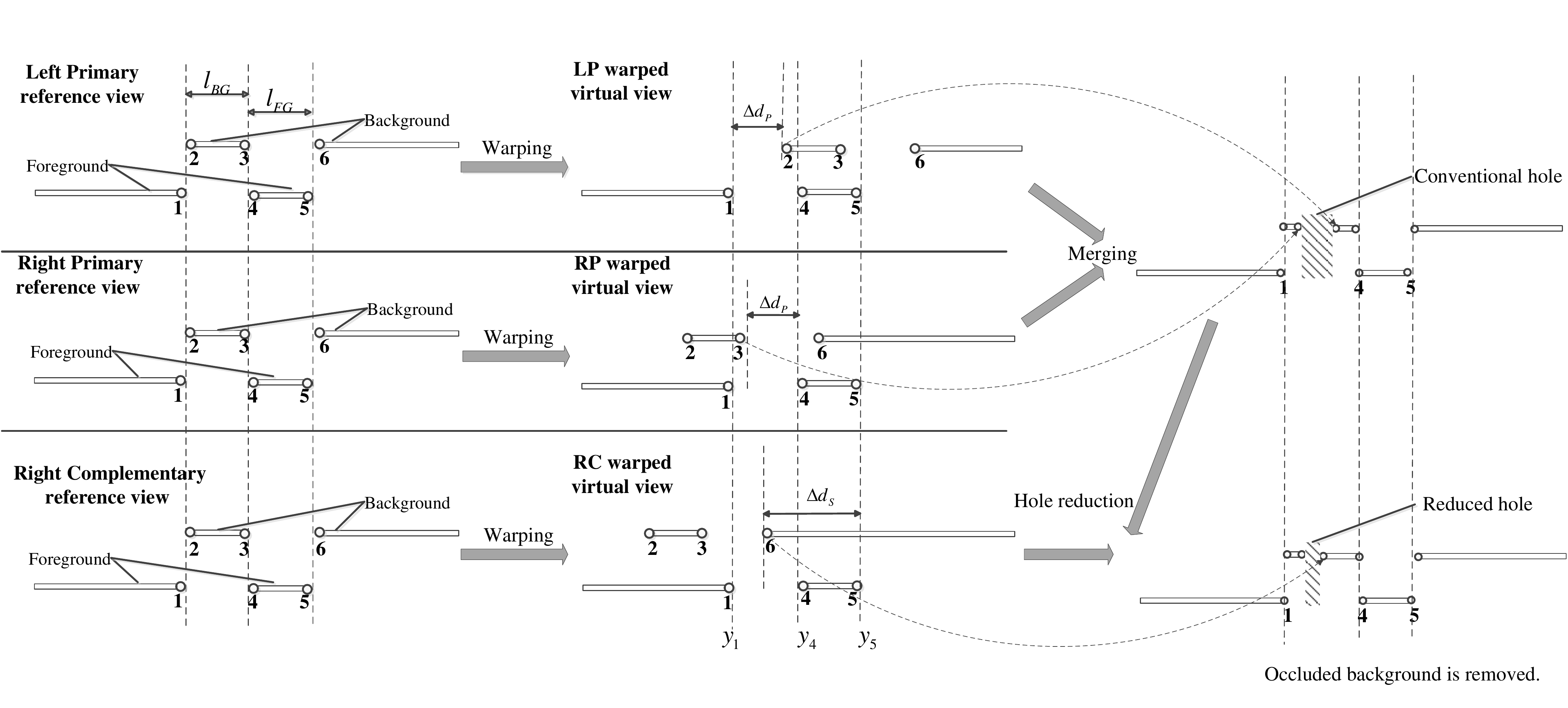}
\caption{View interpolation: illustration of 3D warping process using two primary views and one right complementary view in the F-B-F-B setting.}
\label{fig_vi}
\end{figure*}

\section{View Synthesis Using Multiple Reference Views for Hole Reduction}
\label{MPtheory}
DIBR view synthesis generally consists of three steps: DIBR warping, view merging and hole filling. In the DIBR warping process, pixels in the reference views are projected into the 3D world using the respective depth data and then re-projected to the to-be-synthesized virtual view. The difference of an associated pixel in terms of location in the reference view and the synthesized view is referred to as disparity. In the commonly used parallel camera configuration \cite{MPEGseq}, the vertical disparity is zero and only horizontal shifting of the reference image pixels is involved in the 3D warping process \cite{Tian2009view}. Therefore, for simplicity and without loss of generality, only the horizontal shifting line by line is considered in the following analysis throughout the paper. The horizontal disparity magnitude as the horizontal displacement for each pixel can be obtained by \cite{Tian2009view}

\begin{equation}
\label{equ:Disparity}
d = \frac{f \cdot l}{z}
\end{equation}
where $f$, $l$ and $z$ represent, respectively, the camera focal length, baseline length between the reference view and the target view, and the depth value of the pixel. In view interpolation, there are usually two reference views used for view synthesis. Accordingly, two warped virtual views are obtained with the DIBR warping and then merged together. In this process, certain holes generated in one warped view are compensated by the other warped view. Thus the holes are greatly reduced. On the other hand, view extrapolation generally only employs one reference view and thus no view merging is employed, leading to more holes. At last, hole filling is applied to fill up the holes in the merged view (for view interpolation) or in the warped view (for view extrapolation) as the final virtual view. Details will be further explained in the following Subsections.

\subsection{Multiple Reference Views for View Interpolation: Hole Reduction}
In a typical view interpolation as shown in Fig. \ref{fig_3dv}, two primary reference views are warped to the same viewpoint and fused together to generate a merged view. Some of holes appearing in one warped view may be complemented by the corresponding non-hole region in the other warped view, while the remaining holes in the merged view are the regions that are occluded in both views. These holes may degrade the quality of the virtual view to a great extent in certain scenarios. On the other hand, in the current 3DV and ongoing FTV system, there are generally more than two views available as shown in Fig. \ref{fig_3dv}, and those complementary views may be helpful in improving the quality of the virtual view. Although there are multiple views available on the user side, there has been no thorough investigation on using multiple views for DIBR view synthesis yet. Hence, in the following, we will examine the hole generation mechanism in view interpolation and discuss whether the complementary views are useful in reducing the holes, specifically why and how the complementary views may help reduce holes if so.

Consider a DIBR view synthesis process with three reference views of a scene, i.e., the left primary (LP), right primary (RP) and right complementary (RC) views in a foreground-background-foreground-background (F-B-F-B) setting as shown in Fig. \ref{fig_vi}, where the central viewpoint between the left and right primary views is to be synthesized. To better distinguish the foreground and background, the foreground pixels and background pixels are drawn separately as shown in Fig. \ref{fig_vi}. The foreground boundary points of the views will be warped to the same point since the corresponding foreground boundary points are associated with the same point in the 3D scene. Accordingly, the 3D warping process can be simplified as shown in Fig. \ref{fig_vi}. We denote the locations of the six foreground and background boundary points (from left to right) in the three reference views by $x_i(LP)$, $x_i(RP)$ and $x_i(RC)$, $i=1,2,...,6$, which represent the location of the \textit{i}-th boundary point in the left primary (LP), right primary (RP) and right complementary (RC) reference views, respectively. For the primary and complementary views, the length of the left background segment and the length of the right foreground segment are denoted by $l_{BG}$ and $l_{FG}$, respectively. For simplicity and without loss of generality, the depth values of the two foreground segments are assumed to share the same depth value, and the same for the two background segments. Similar results can be obtained with varying depth values for the foreground and background segments, respectively. The magnitudes of horizontal disparity differences between foreground and background with respect to the virtual viewpoint for the primary and complementary views are denoted by $\Delta d_p$ and $\Delta d_s$, respectively.

\begin{figure*}[tbp]
\centering
\includegraphics[width=1\hsize]{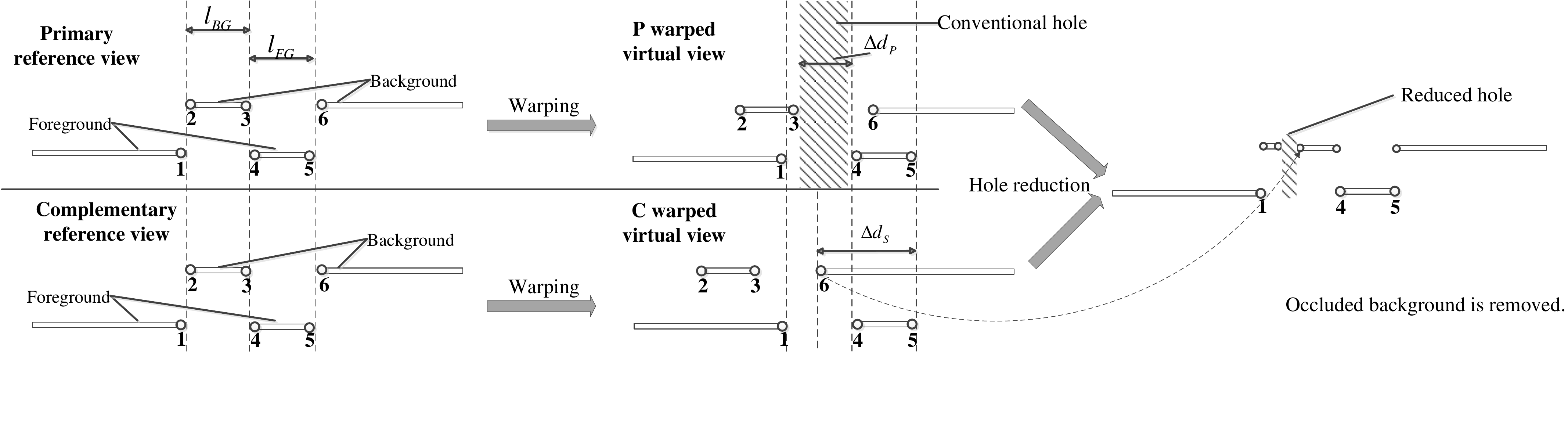}
\caption{View extrapolation: illustration of 3D warping process using one primary and one complementary view in the B-F-B setting.}
\label{fig_ve}
\end{figure*}

As shown in Fig. \ref{fig_vi}, the 18 boundary points $x_i$ in the three reference views (6 in each) are mapped to $y_i$ in the warped virtual views, respectively. For example, $y_2(LP)$ is the corresponding warped point of $x_2(LP)$, whereas $y_6(RC)$ in the RC warped virtual view corresponds to $x_6(RC)$ in the RC reference view. As the foreground boundary points $x_1(LP)$, $x_1(RP)$, $x_1(RC)$ in the three reference views are associated with the same point in the 3D scene, they are expected to be mapped to the same point in the virtual view, which suggests $y_1(LP) = y_1(RP) = y_1(RC) \equiv y_1$. Such three to one point mapping also applies for the other two foreground boundary points $x_4$ and $x_5$, which means $y_4(LP) = y_4(RP) = y_4(RC) \equiv y_4$ and $y_5(LP) = y_5(RP) = y_5(RC) \equiv y_5$. Since the depth values of each segment are assumed to be the same, the length of each segment will remain unchanged when warped into the virtual view. As indicated in Fig. \ref{fig_vi}, the locations of other warped boundary points in the virtual view can be obtained based on their horizontal disparity. As a summary, the locations of the warped boundary points with reference to $y_1$ in the virtual view follow

\begin{equation}
\begin{split}
& \left\{ \begin{array}{l}
y_2(LP) = y_1 + 1 + \Delta d_p \\
y_3(LP) = y_1 + l_{BG} + \Delta d_p \\
y_6(LP) = y_1 + l_{BG} + l_{FG} + 1 + \Delta d_p
\end{array} \right. \\
& \left\{ \begin{array}{l}
y_2(RP) = y_1 + 1 - \Delta d_p \\
y_3(RP) = y_1 + l_{BG} - \Delta d_p \\
y_6(RP) = y_1 + l_{BG} + l_{FG} + 1 - \Delta d_p
\end{array} \right. \\
& \left\{ \begin{array}{l}
y_2(RC) = y_1 + 1 - \Delta d_s \\
y_3(RC) = y_1 + l_{BG} - \Delta d_s \\
y_6(RC) = y_1 + l_{BG} + l_{FG} + 1 - \Delta d_s
\end{array} \right. \\
\end{split}
\end{equation}
while the locations $y_4$ and $y_5$ follow
\begin{equation}
\left\{ \begin{array}{l}
y_4 = y_1 + l_{BG} + 1 \\
y_5 = y_1 + l_{BG} + l_{FG} \\
\end{array} \right.
\end{equation}
where ``+1'' refers to the right neighboring pixel just by one unit (note that $y_1$ indicates the location of the foreground boundary point corresponding to $x_1$ in the reference view while ``+1'' refers to the starting point of the neighboring background segment in the right as shown in Fig. \ref{fig_vi}).

When view merging is applied, two warped images of the two primary reference views will be blended and some of the holes will be filled by the other warped image. As can be seen in Fig. \ref{fig_vi}, a hole from $max(y_3(RP)+1, y_1+1)$ to $min(y_2(LP)-1, y_4-1, y_6(RP)-1)$, which is known as the conventional hole rendered based on two reference views, appears in the blended view if the latter is larger. When the right complementary view is taken into consideration, the background segment starting from $y_6(RC)$ may fill part or whole of the hole if $y_6(RC)$ is at the left side of the right endpoint of the hole, which diminishes the hole size. That is to say, the hole length with the complementary right view will be reduced to the range of $max(y_3(RP)+1, y_1+1)$ to $min(y_2(LP)-1, y_4-1, y_6(RC)-1)$. When $y_6(RC)-1$ becomes smaller than $max(y_3(RP)+1, y_1+1)$, the hole will be completely eliminated. Therefore, incorporating the complementary right view into the view synthesis based view interpolation is helpful to diminish holes in the merged view.

Likewise, in the case of background-foreground- background-foreground (B-F-B-F) setting, it can be shown that the left complementary view can be used for hole reduction. Therefore, in the current 3DV system with three (or more) views coded and transmitted to the decoder as shown in Fig. \ref{fig_3dv}, the complementary view(s) other than the two primary views generally used in view synthesis can be employed to improve the synthesis quality. In a more general case with more depth discontinuities, it can be shown that the complementary views can jointly help reduce holes when an F-B-F-B setting and a B-F-B-F setting are present in the scene. Further, the FTV system may achieve significantly better performance in view synthesis using multiple views (more than two views). Therefore, we propose a new view synthesis framework to exploit multiple views in synthesizing all the virtual views.

%\linespread{1.7}
\subsection{Multiple Reference Views for View Extrapolation: Hole Reduction}
View extrapolation has been investigated originally in the application of using just one view of texture video and its corresponding depth video to synthesize a new virtual view. Different from view interpolation using the left and right views together to produce an in-between virtual view, view extrapolation only employs one reference view. It results in more holes in a virtual view, due to the absence of the other view in the opposite direction that can complement to fill the holes introduced in the case of single depth discontinuity. Among the holes, some may be diminished if complementary views in the same direction are exploited.

Taking for example that using the left view in Fig. \ref{fig_3dv} to synthesize views in the left beyond the covering view range, we study whether the central view can be exploited to reduce holes in the synthesized views. In such a setting as shown in Fig. \ref{fig_3dv}, the left view is treated as the primary view and the central view is regarded as a complementary view. In view interpolation, warped views from two primary reference views are merged to produce the synthesized view, and thus the holes appeared in the merged view present necessarily in both of the warped views. In other words, such holes appear when only using one primary view for view extrapolation. Likewise, information from the complementary view may also reduce the hole sizes in view extrapolation under the same condition as in Subsection A for view interpolation.

Compared to the case of view interpolation, the complementary view tends to be more helpful in reducing the holes in view extrapolation. In view interpolation, the holes generated in one warped view may be complemented by the other view, which leads to no holes in the case of single depth discontinuity and background-foreground-background (B-F-B) discontinuity \cite{MyJour_TBC}. However, in the view extrapolation, only one primary view is employed and hence holes still remain in such cases. In the following we will show that the complementary view may help reduce holes in such cases.

Consider the case of B-F-B discontinuity as shown in Fig. \ref{fig_ve}. For simplicity, we also make the assumptions and the notations of view interpolation in Subsection A, where $x_i(P)$ and $x_i(C)$, $i=1,2,3,4$, represent the locations of the $i$-th boundary point in the primary and complementary reference views, respectively, and $y_i(P)$ and $y_i(C)$ their corresponding mapped pixels in the primary and complementary warped virtual views, respectively.

Similarly as we discussed in Subsection A, $x_2(P)$ and $x_2(C)$ will be warped to the same point in the virtual view, and the same for $x_3(P)$ and $x_3(C)$, which means $y_2(P)=y_2(C) \equiv y_2$ and $y_3(P)=y_3(C) \equiv y_3$. The locations of the other boundary points can be obtained with reference to $y_2$ as shown in the following.

\begin{equation}
\begin{split}
& \left\{ \begin{array}{l}
y_1(P) = y_2 - 1 + \Delta d_p \\
y_4(P) = y_2 + l_{FG} - \Delta d_p
\end{array} \right. \\
& \left\{ \begin{array}{l}
y_1(C) = y_2 - 1 - \Delta d_s \\
y_4(C) = y_2 + l_{FG} - \Delta d_s
\end{array} \right. \\
\end{split}
\end{equation}

As indicated in Fig. \ref{fig_ve}, a hole from $y_1+1$ to $min(y_4(P)-1, y_2-1)$ appears in the virtual view in view extrapolation if the latter is larger. Since the baseline distance between the virtual view and the complementary view is larger than that between the virtual view and the primary view, $\Delta d_s$ is larger than $\Delta d_p$ according to (1). Hence, when considering the complementary view in view extrapolation, the background segment from $y_4(C)$ to $min(y_4(P)-1, y_2-1)$ may fill the hole. It will be completely eliminated when $y_4(C)$ becomes smaller than $y_1(P)+1$. Likewise, when using view extrapolation to generate the virtual views on the right side of the right view, the central view can also be useful under certain circumstances. More generally, when we extrapolate virtual views using one specific view, the other available views may also be useful in addition to the closest reference one. For example, in the case of three views as in Fig. \ref{fig_3dv}, the central view and the right view can both serve as complementary reference views to help synthesize views on the left side of the left view. Therefore, as suggested in Subsection A, in the 3DV and FTV systems, a new view synthesis framework to use multiple reference views to collectively synthesize all the virtual views would achieve higher quality than conventionally using one reference in view extrapolation or two references in view interpolation.

\begin{figure}[tbp]
\centering
\includegraphics[width=1\hsize]{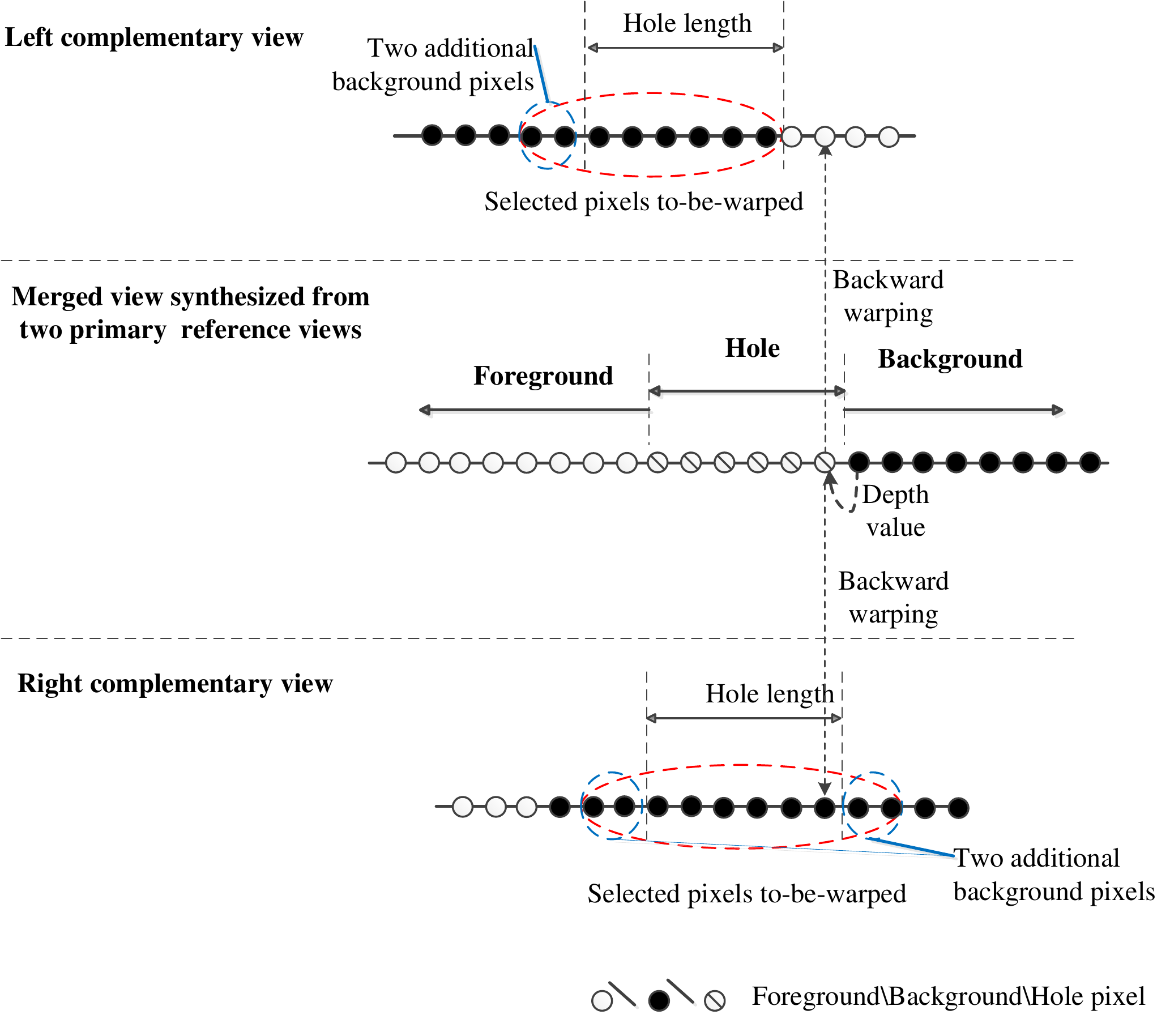}
\caption{An illustration of the selective warping scheme using the left and right complementary views.}
\label{fig_select}
\end{figure}

\section{Selective Warping of Complementary Views for Hole Reduction}
\label{selectivewarp}
In the new framework of exploiting all available reference views for synthesized views of better quality, a full warping of any complementary view to assist in reducing the holes would greatly aggravates the computational complexity in the DIBR view synthesis, which is also a waste as only a small portion of the warped view is useful for hole reduction. To lower the complexity while still warping those pixels useful for hole reduction, a selective warping scheme is highly desired, in which the key is to locate the relevant pixels in the complementary views for warping.

In view that holes are the background regions occluded by the foreground objects in the reference views but being visible in a synthesized view, only the background pixels of the complementary views may be projected into the hole regions. Therefore, the relevant pixels in the complementary views can be located by backward warping the hole locations with correct background depth values that are unknown yet and to be estimated. To further lower the rendering complexity, only the depth value of one edge pixel of a hole is estimated and then backward warped as the other relevant pixels can be easily located by searching its neighboring background pixels along the same direction in the complementary views.

An illustration of the proposed selective warping from complementary views is depicted in Fig. \ref{fig_select}. First, the background depth value in a hole is estimated. As in our previous work \cite{MyJour_TBC}, we have shown that the occluded background information can be used to determine whether the information around the hole belongs to background. After determining the background information around the hole, the (left or right) edge pixel of the hole which is adjacent to the background is then assigned by the depth value of its most adjacent background pixel since the depth map is generally flat in the background regions. If no background information is available around the hole, the depth value of the closest occluded background is treated as the depth value of the hole. With the estimated depth value, a background pixel in a complementary view is located by backward warping the hole edge pixel. Accordingly, its neighboring background pixels of the hole length in the same direction  with a few more additional background pixels (e.g., two in Fig. \ref{fig_select}) in the neighborhood are the selected pixels for warping. In Fig. \ref{fig_select}, two additional background pixels are included to accommodate some possible depth variations, which may lead to slight changes in warping locations.

From the above, it can be seen that the proposed selective warping process only warps a small number of the relevant background pixels in the complementary views for hole reduction, thus substantially saving computation which will be validated in the following section. Another benefit from the selective scheme is better quality of the pixels warped in the hole, as the warping of irrelevant pixels in the complementary reference view can be skipped which may be associated with inaccurate or erroneous depth values due to the state-of-the-art and relatively poor depth generation techniques.

\section{Experimental Results}
\label{experiment}
Simulations were performed based on the View Synthesis Reference Software (VSRS) Version 3.5 \cite{VSRS}. Four test sequences \textit{Ballet}, \textit{Breakdancers} \cite{Ballet}, \textit{UndoDancer} \cite{CTC}, \textit{Shark} and \textit{Bee} \cite{FTV_seq} are considered to evaluate the proposed view synthesis schemes with multiple reference views. Note that \textit{Shark} is an FTV sequence which is different from that used in the 3DV system. 100 frames are synthesized for \textit{Ballet}, \textit{Breakdancers}, and 250 frames are synthesized for \textit{UndoDancer}. \textit{Shark} is a sequence of 100 frames with the first and last few frames containing no object but water which results in no holes in the synthesized view. Hence, only 40 frames from frame 46 to frame 85 of \textit{Shark} are used to evaluate the performance. On the other hand, \textit{Bee} is a still image of 185 views. Among the test sequences, \textit{Ballet} and \textit{Breakdancers} are of resolution 1024*768, while  \textit{UndoDancer}, \textit{Shark} and \textit{Bee} are of resolution 1920*1088. Experiments were conducted on a personal desktop computer of a CPU core i5 (3.2GHz) and 8 GB memory.

In the following, the conventional view interpolation result based on the two primary views (PV) is denoted by 2PV, while the 2PV plus the full warping and the selective warping (SW) of two complementary views (CV) are referred to as 2PV+2CV and 2PV+SW\_2CV, respectively. The conventional view extrapolation result based on one primary view (PV) is denoted by PV, while the PV plus the full warping and the selective warping (SW) of one complementary view (CV) are referred to as PV+CV and PV+SW\_CV, respectively. Note that the conventional hole filling process based on image inpainting \cite{SPIC_FTV,VSRS} is disabled in the current setting as the objective of the experiments is to evaluate the hole size in terms of number of pixels in the holes by different approaches. Also a margin of 60 pixels to the image boundaries is not counted as the holes since the holes in this area are mainly due to the difference in the capture angle or range of each camera instead of disocclusion.

\begin{table}[t]
   \caption{\textsc{View interpolation: test sequences and settings.}}
   \centering
   \tabcolsep=4.8pt
   \begin{tabular}{@{}c c c c@{}}%|p{4em}|p{3em}|p{3.5em}|p{3.5em}|p{3.5em}|p{3.5em}|
	\hline
	Sequence & Target View & Primary Views & Complementary Views\\
	\hline
	\textit{Ballet} & View 4	& View 3, View 5 & View 1, View 7\\
	\hline
	\textit{Breakdancers} & View 4	& View 3, View 5 & View 1, View 7\\
	\hline
	\textit{Shark} & View 75	& View 60, View 90 & View 30, View 120\\
	\hline
	\textit{Bee} & View 100	& View 75, View 125	& View 50, View 150 \\
	\hline
	\end{tabular}
   \label{vi_ts}
\end{table}

\begin{table}[t]
   \caption{\textsc{View interpolation: hole size comparison by different approaches in terms of average number of pixels in holes (per frame).}}\label{vi_size}
   \centering
   \begin{tabular}{c | c c c | c c}%|p{4em}|p{3em}|p{3.5em}|p{3.5em}|p{3.5em}|p{3.5em}|
	\hline
	 \multirow{3}{*}{Sequence} & \multicolumn{3}{c|}{Hole Size}  & \multicolumn{2}{c}{Hole Reduction (\%)} \\ 
	 \cline{2-6}
	 & \multirow{2}{*}{2PV} & 2PV+ & 2PV+ & 2PV+ & 2PV+  \\
	  & & 2CV & SW\_2CV &  2CV & SW\_2CV \\ 
	\hline
	\textit{Ballet} & 877	& 96	& 210	& 89.02	& 76.09\\
	\hline
	\textit{Breakdancers} & 422	& 27	& 235	& 93.63	& 44.37 \\
	\hline
	\textit{Shark} & 1715	& 660	& 818	& 61.49	& 52.27 \\
	\hline
	\textit{Bee} & 69563	& 37842	& 46638	& 45.6	& 32.96 \\
	\hline
	\end{tabular}
   
\end{table}
\begin{table}[t]
   \caption{\textsc{View interpolation: PSNR comparison for the reduced hole pixels by 2PV+2CV and 2PV+SW\_2CV. (The difference between the two columns by the 2PV+Inpainting is due to the reduced number of filled pixels by 2PV+SW\_2CV compared with 2PV+2CV.)}}\label{vi_psnr}
   \centering
   \begin{tabular}{c | @{\hskip 0.18in}  c @{\hskip 0.25in}  c @{\hskip 0.18in}  |  @{\hskip 0.18in} c c }%|p{4em}|p{3em}|p{3.5em}|p{3.5em}|p{3.5em}|p{3.5em}|
	\hline
	 \multirow{3}{*}{Sequence} & \multicolumn{4}{c}{PSNR of Filled Pixels by Two Testing Approaches (dB)} \\ %Respectively
	 \cline{2-5}
	 & 2PV+ & 2PV+ & 2PV+ & 2PV+  \\
	  & Inpainting & 2CV & Inpainting & SW\_2CV \\ 
	\hline
	\textit{Ballet} & 23.28	& 25.23	& 22.94	& 27.47\\
	\hline
	\textit{Breakdancers} & 21.17	& 14.94	& 20.74	& 18.54 \\
	\hline
	\textit{Shark} & 21.09	& 29.01	& 21.3	& 30.84 \\
	\hline
	\textit{Bee} & 15.76 & 19.53 & 16.07 & 19.86\\
	\hline
	\end{tabular}
   
\end{table}

\begin{table}[t]
   \caption{\textsc{View interpolation: complexity comparison by different approaches in terms of average rendering time in sec (per frame).}}
   \centering
   \begin{tabular}{c c c c }%|p{4em}|p{3em}|p{3.5em}|p{3.5em}|p{3.5em}|p{3.5em}|
	\hline
	Sequence & 2PV & 2PV+2CV & 2PV+SW\_2CV\\
	\hline
	\textit{Ballet} & 0.41	& 0.792	& 0.453\\
	\hline
	\textit{Breakdancers}& 0.409	& 0.805	& 0.451\\
	\hline
	\textit{Shark}& 1.048	& 2.042	& 1.164\\
	\hline
	\textit{Bee} & 1.076	& 2.059	& 1.232 \\
	\hline
	\end{tabular}
   \label{vi_complexity}
\end{table}

\subsection{View Interpolation}
The test sequences along with the specification of the reference views and target virtual views used for view interpolation are shown in Table \ref{vi_ts}. Table \ref{vi_size} shows the hole size in the merged view per frame by using 2PV, 2PV+2CV and 2PV+SW\_2CV, as well as the hole reduction percentages by 2PV+2CV and 2PV+SW\_2CV over 2PV, respectively. It can be seen that most of the holes (about 90\% for \textit{Ballet} and \textit{Breakdancers} and 60\% for \textit{Shark}) in the conventional 2PV view synthesis can be filled by warping the complementary views. Though the hole reduction by the proposed selective warping scheme is generally less than that by the full warping, the quality of the filled pixels in the selective warping scheme is found to be better than that of full warping as shown in Table \ref{vi_psnr} with the reason mentioned in the preceding section. For example, for \textit{Ballet} sequence, the PSNR value for the pixels filled in the hole by 2PV+SW\_2CV is more than 2 dB higher than that by 2PV+2CV (27.5 dB versus 25.2 dB). For \textit{Breakdancers} sequence, the hole reduction percentage by the selective warping scheme appears to be not as good as those for the \textit{Ballet} and \textit{Shark} sequences. It is mainly due to the poor depth quality of the \textit{Breakdancers} sequence which significantly affects the warping performance.

\begin{figure}[t]
  \centering
   \centerline{\subfigure[\textit{Ballet}]{\includegraphics[width=0.9\hsize]{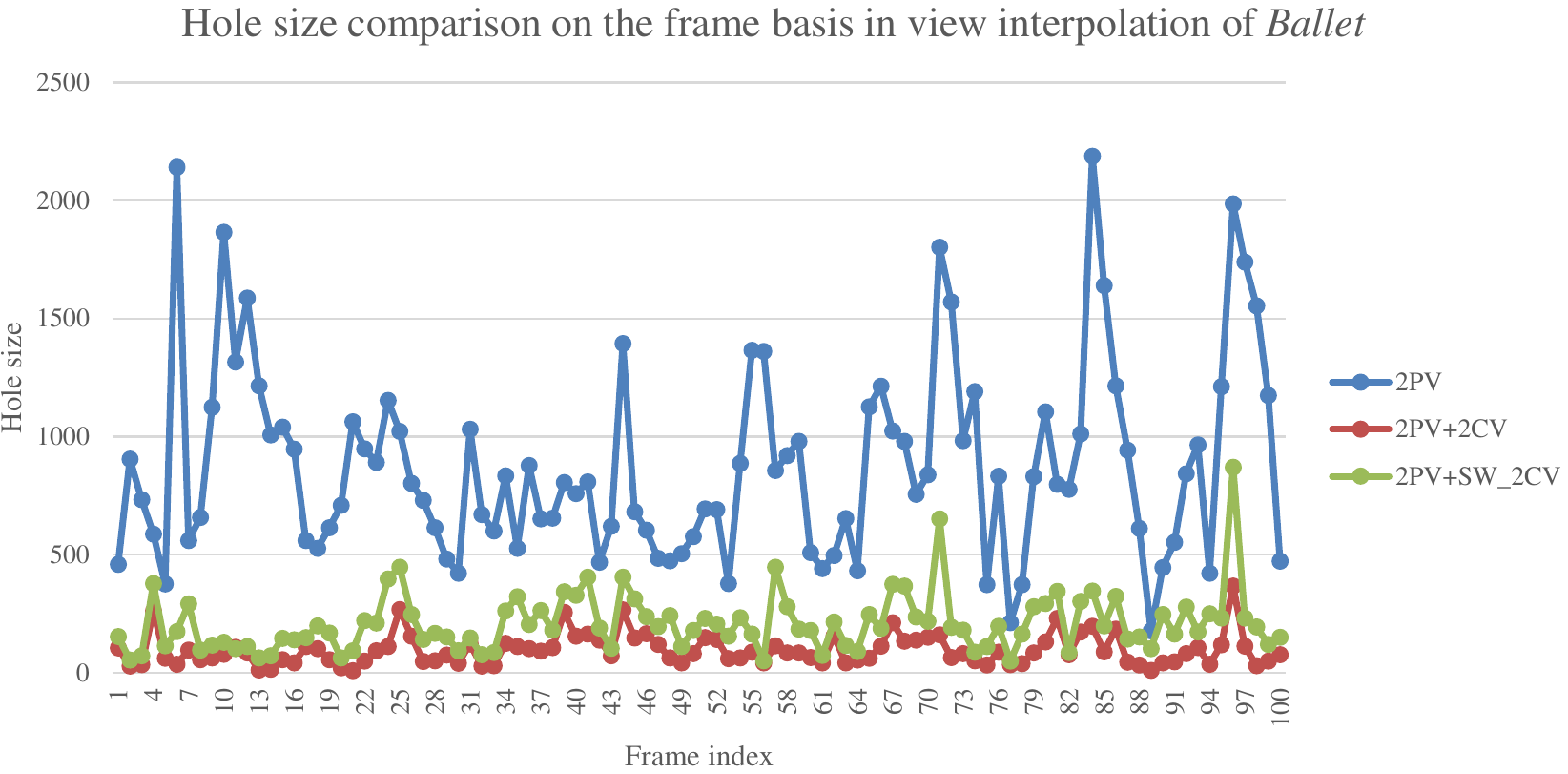}}   }
   %\vspace{0.2cm}
   \centerline{\subfigure[\textit{Shark}]{\includegraphics[width=0.9\hsize]{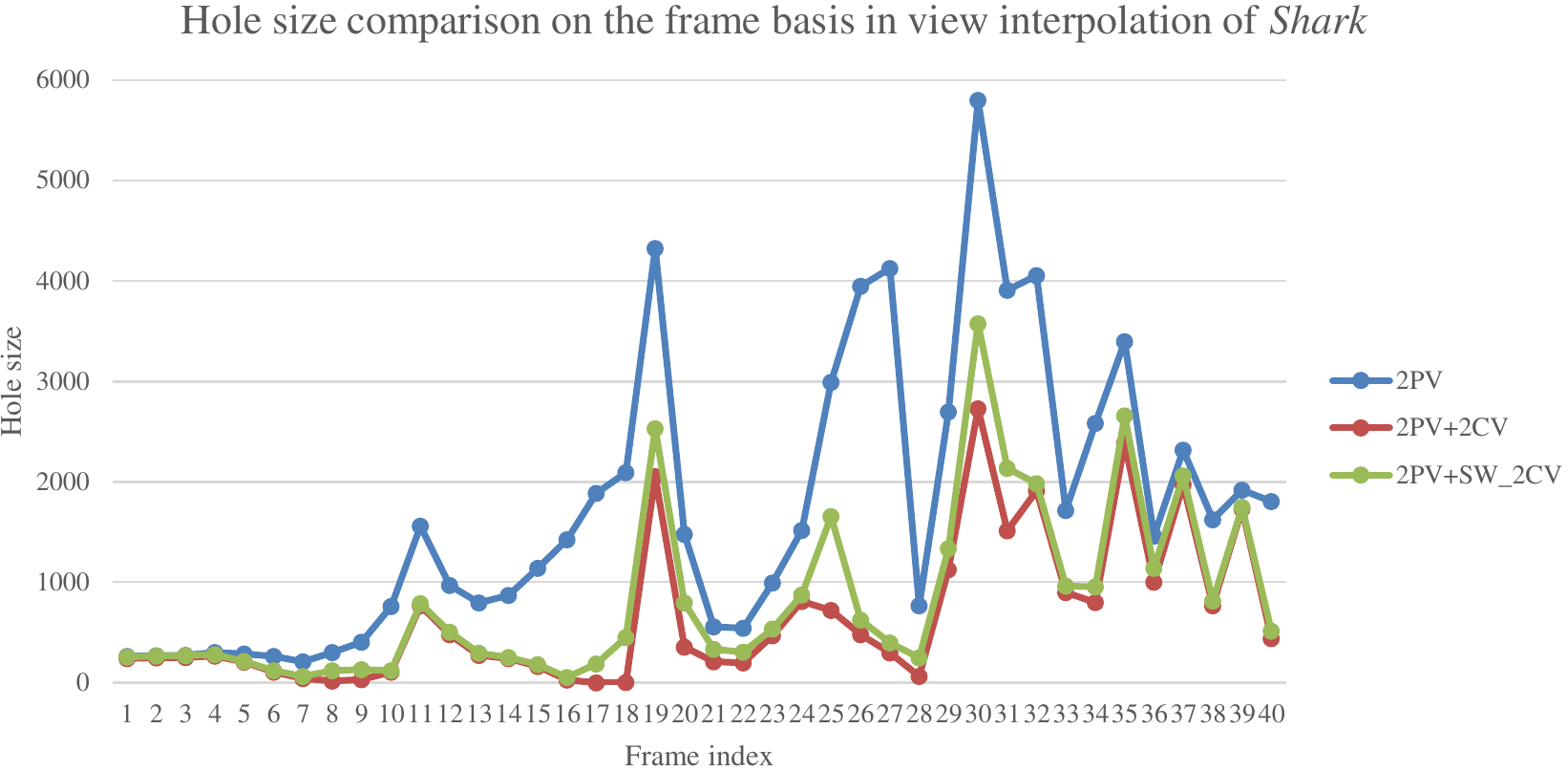}}   }
   \caption{Hole size comparison on the frame basis in view interpolation.}
   \label{fig_vi_size_perframe}
\end{figure}

\begin{figure}[t]
  \centering
   \centerline{\subfigure[2PV+Inpainting and 2PV+2CV]{\includegraphics[width=0.9\hsize]{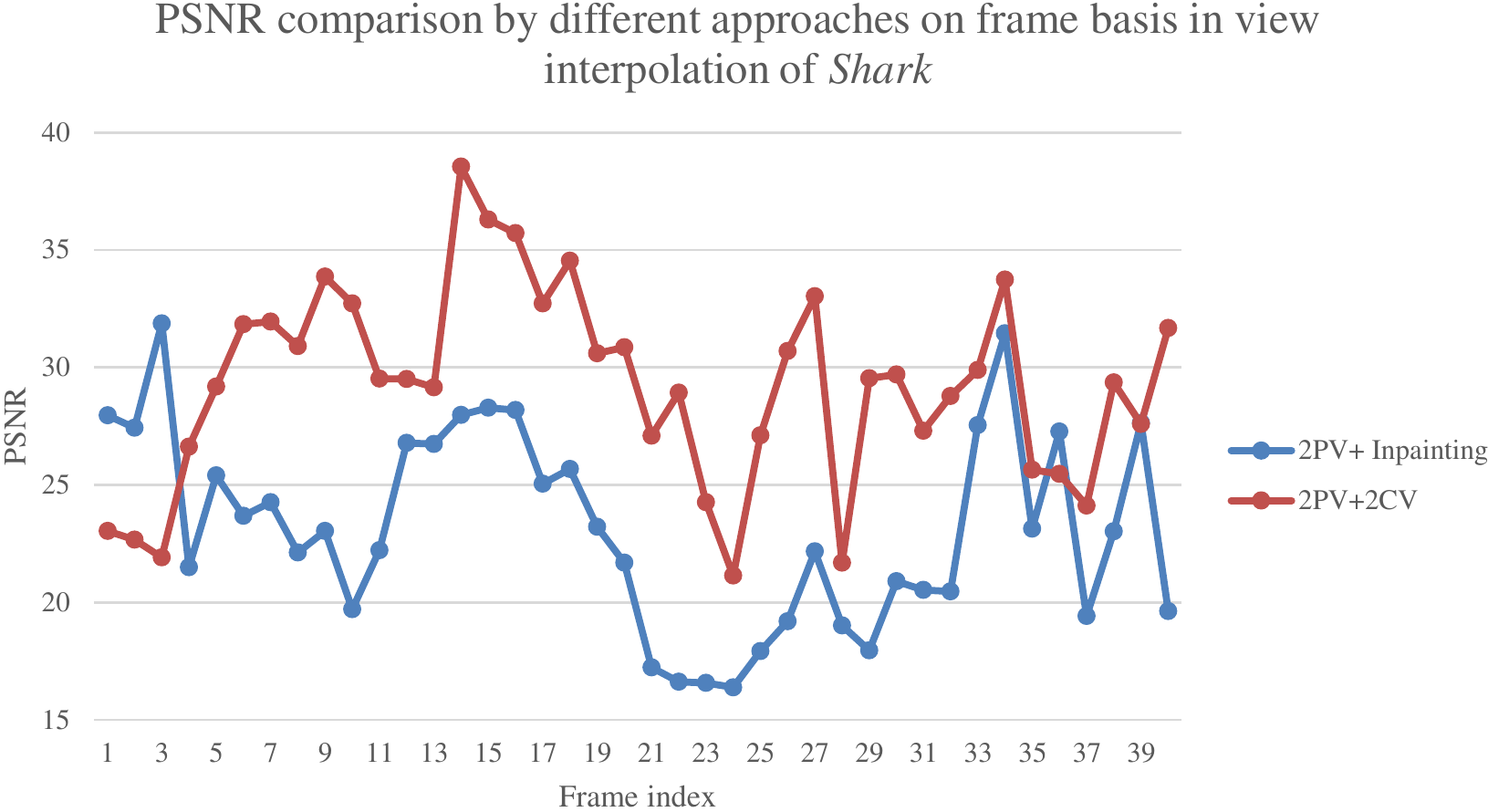}}   }
   \vspace{-0.2cm}
   \centerline{\subfigure[2PV+Inpainting and 2PV+SW\_2CV]{\includegraphics[width=0.9\hsize]{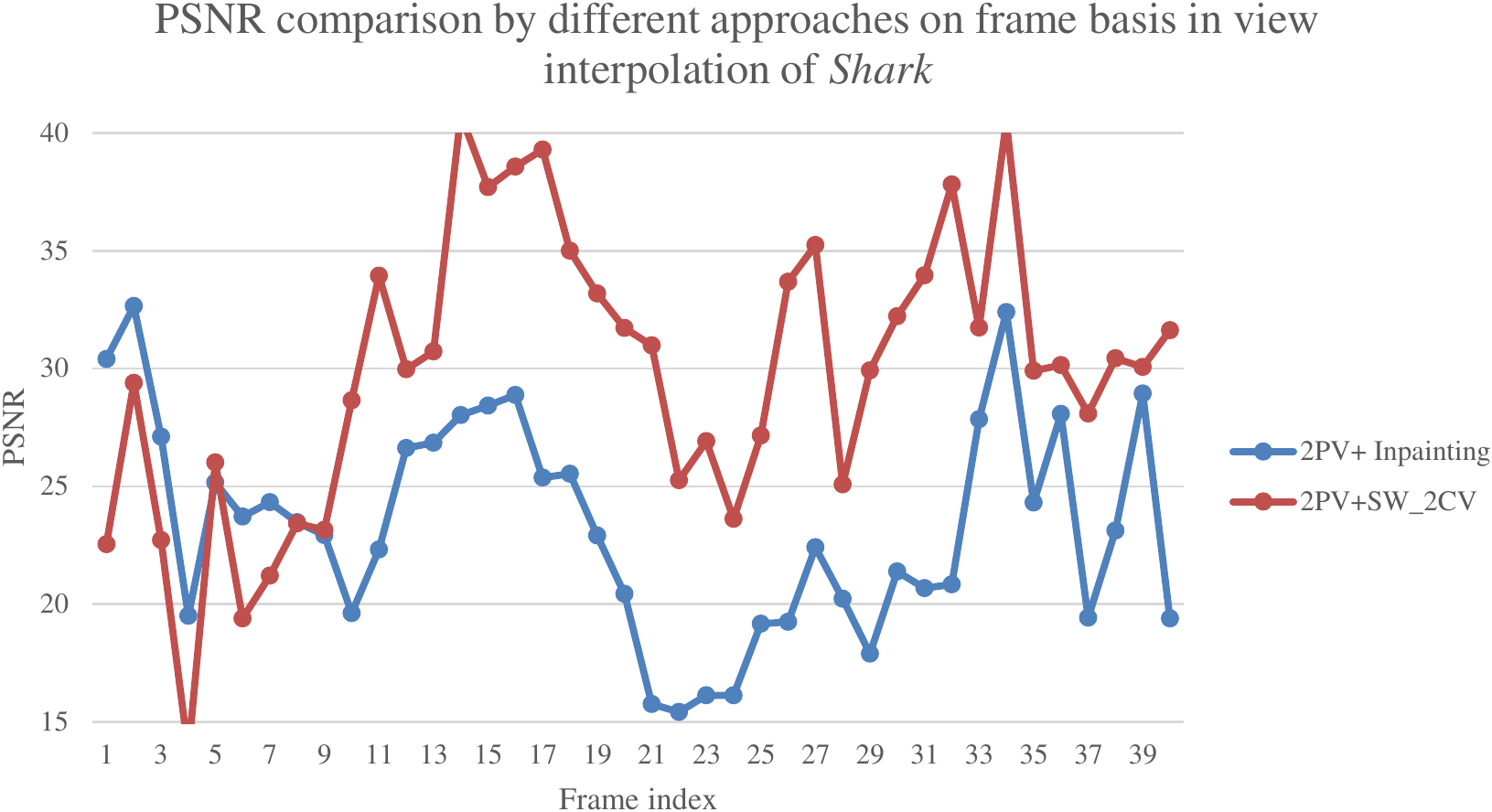}}   }
   \vspace{-0.2cm}
   \caption{PSNR comparison by different approaches on the frame basis in view interpolation for \textit{Shark}.}
   \label{fig_vi_psnr_perframe}
\end{figure}

Table \ref{vi_psnr} also shows quality comparison of the filled pixels by the proposed methods and image inpainting with respect to the original image. It can be seen that among the four testing sequences except \textit{Breakdancers}, the quality of the filled pixels is generally higher than those filled by the image inpainting method in VSRS. The reason of the inferior filling quality for the \textit{Breakdancers} is still the poor quality of its depth map mentioned above. On the contrary, \textit{Shark} is a video sequence generated by computer graphics where the quality of its depth maps is very high. Therefore, quality of the filled pixels by our proposed method are much higher than those filled by image inpainting. Note that the small differences of the PSNR values of the pixels filled by the image inpainting method (e.g., the two columns by 2PV +inpainting in Table \ref{vi_psnr}) are due to that the hole pixels used in the comparisons  are different (which are determined by the reduced holes using the 2PV+2CV and 2PV+SW\_2CV, respectively). To further demonstrate the robustness of our proposed method, Fig. \ref{fig_vi_size_perframe} compares the hole sizes using different methods in each frame of \textit{Ballet} and \textit{Shark}, respectively. It can be seen that our proposed method consistently reduces the hole size throughout all the testing frames in the video sequences. We can also see that the performance (in terms of hole size reduction) on \textit{Ballet} is better than \textit{Shark}. It is mostly due to that our proposed methods work better on images with more depth discontinuities as described in Section \ref{MPtheory}. For \textit{Ballet}, it is a person with arms and legs that leads to more depth discontinuities, compared with \textit{Shark}. Consequently the hole size reduced by our proposed methods for \textit{Ballet} is larger than that of \textit{Shark}. PSNR comparison of our proposed methods against image inpainting for each frame of \textit{Shark} is shown in Fig. \ref{fig_vi_psnr_perframe}. It can be seen that our proposed method generally shows higher quality than image inpainting.

%\FloatBarrier

\begin{figure}[!htbp] 
  \addtocounter{subfigure}{-1}
  %\vspace{-0.1cm}
   \centerline{\subfigure{\includegraphics[width=1\hsize]{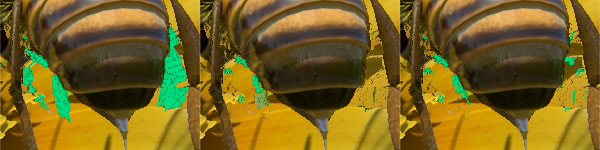}}   }
   \vspace{-0.25cm}
   \addtocounter{subfigure}{-1}
   \centerline{\subfigure{\includegraphics[width=1\hsize]{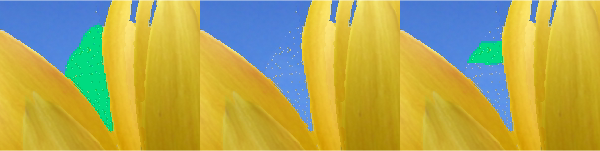}}   }
   \vspace{-0.25cm}
   \centerline{\subfigure[\textit{Bee}]{\includegraphics[width=1\hsize]{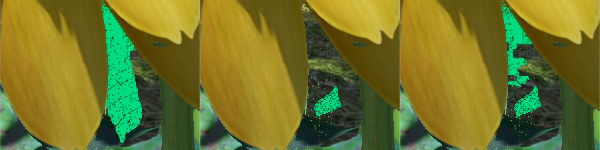}}   }  
   
   \caption{Snapshots of some synthesized views. The pictures from left to right denote the synthesized views using 2PV, 2PV+2CV and 2PV+SW\_2CV, respectively, with holes in green.}
   \label{fig_vi_snapshots}
\end{figure}

Fig. \ref{fig_vi_snapshots} shows some snapshots of the synthesized images by the three approaches. The pictures in each row from left to right demonstrate the synthesized views using 2PV, 2PV+2CV and 2PV+SW\_2CV, respectively, with holes indicated in green. It can be clearly seen that most of the holes can be reduced or eliminated by including the warping of the complementary views in the 2PV+2CV and 2PV+SW\_2CV schemes. Also, from the pictures of \textit{Breakdancers} in the second row, it is visible that the quality of the pixels filled in the big hole by the selective warping approach appears to be better than that by the full warping approach (that is, the pixels filled by the SW\_2CV look to be much closer to background pixels). Note that the contour noise around the hole is due to the inaccurate depth values near the object boundary and the color bleeding effect of the texture images, known as boundary artifact or boundary noise in view synthesis \cite{zhu_boundary}.

To compare the complexity, the rendering time per frame by each scheme is tabulated in Table \ref{vi_complexity}. It can be seen that the rendering time by 2PV+2CV is about 2 times of that by 2PV as expected, while the selective warping scheme only increases about 12\% rendering time over the 2PV by reducing the complexity of fully warping two complementary views by  88\% in average. Note that the time of loading frames of different views and producing output frames is ignored since in the practical applications all the views would be loaded together to produce outputs of different views. 

Moreover, to illustrate the effectiveness of our proposed framework in synthesizing a virtual view using all available views, we further conduct an experiment by using another two complementary views (i.e., four complementary views and two primary views) for view synthesis. The results of the hole sizes are tabulated in Table \ref{vi_4_hole}. Due to the limited number of views available, only \textit{Shark}  and \textit{Bee} is used in the test. For \textit{Shark}, View 9 (which is the smallest view number provided) and View 150 are used in addition to the setting in 2PV+2CV while for \textit{Bee}, View 50 and View 150 are used. It can be seen from Tables \ref{vi_size} and \ref{vi_4_hole} that around 10\% hole pixels are reduced additionally by 2PV+4CV (or 2PV+SW\_4CV) compared against 2PV+2CV (or 2PV+SW\_2CV) using two farther complementary views. The PSNR comparison using different approaches are shown in Table \ref{vi_4_psnr}, where the PSNRs of all the hole pixels are used. It can be seen that by using extra complementary views, the quality is further improved. If only considering the hole pixels filled by 2PV+ 4CV or 2PV+ SW\_4CV, the quality can be much better than those filled by 2PV+inpainting. Taking \textit{Shark} for example, the quality of hole pixels filled by 2PV+4CV reaches 28.48 dB while only 20.79 dB for 2PV+inpainting, and the quality of hole pixels filled by 2PV+SW\_4CV reaches 30.45 dB while 21.14 dB for 2PV+inpainting. The complexity comparison of using additional complementary views is shown in Table \ref{vi_4_complexity}, where for the selective warping approach, around 25.4\% time is increased. Compared with the time increase shown in Table IV, the time increase is doubled, which can be expected since warping additional complementary views uses the same approach as warping the first pair of complementary views.

\begin{table}[t]
   \caption{\textsc{View interpolation: hole size comparison by different approaches (using another pair of views) in terms of average number of pixels in holes (per frame).}}\label{vi_4_hole}
   \centering
   \begin{tabular}{c | c c c | c c}%|p{4em}|p{3em}|p{3.5em}|p{3.5em}|p{3.5em}|p{3.5em}|
	\hline
	 \multirow{3}{*}{Sequence} & \multicolumn{3}{c|}{Hole Size}  & \multicolumn{2}{c}{Hole Reduction (\%)} \\ 
	 \cline{2-6}
	 & \multirow{2}{*}{2PV} & 2PV+ & 2PV+ & 2PV+ & 2PV+  \\
	  & & 4CV & SW\_4CV &  4CV & SW\_4CV \\ 
	\hline
	\textit{Shark}& 1715	& 385	& 570	& 77.54	& 66.77 \\
	\hline
	\textit{Bee} & 69563	& 32757	& 42457	& 52.91	& 38.97 \\
	\hline
	\end{tabular}
   
\end{table}

\begin{table}[t]
\tabcolsep=3.8pt
   \caption{\textsc{View interpolation: PSNR comparison} (dB) \textsc{for the hole pixels filled by different approaches.}} \label{vi_4_psnr}
   \centering   
   \begin{tabular}{c  c  c  c c c }%!{\vrule width 1.6pt}
	\hline
%	 \multirow{3}{*}{Sequence} & \multicolumn{5}{c}{PSNR of Filled Pixels (dB)} \\ 
%	 \cline{2-6}
	 \multirow{3}{*}{Sequence}& 2PV+ & 2PV+ & 2PV+ & 2PV+ & 2PV+ \\
	  & &2CV+ & SW\_2CV+ & 4CV+ & SW\_4CV+  \\
	  & Inpainting & Inpainting & Inpainting  & Inpainting & Inpainting \\ 
	\hline
	\textit{Shark} & 19.97	& 21.97	& 21.95	& 22.93 & 22.87 \\
	\hline
	\textit{Bee} & 16.32 & 18.01 & 17.65 & 18.82 & 18.39 \\
	\hline
	\end{tabular}
%\end{table}
%
%\begin{table}[t]
\bigskip
   \caption{\textsc{View interpolation: complexity comparison by different approaches (using another pair of views) in terms of average rendering time in sec (per frame).}} \label{vi_4_complexity}
   \centering  
   \begin{tabular}{c c c c }%|p{4em}|p{3em}|p{3.5em}|p{3.5em}|p{3.5em}|p{3.5em}|
	\hline
	Sequence & 2PV & 2PV+4CV & 2PV+SW\_4CV\\
	\hline
	\textit{Shark}& 1.030	& 2.592	& 1.247\\
	\hline
	\textit{Bee} & 1.045	& 2.980	& 1.357 \\
	\hline
	\end{tabular}
\end{table}

\begin{figure}[!htbp]
  \centering
  \addtocounter{subfigure}{-1}
   \centerline{\subfigure{\includegraphics[width=1\hsize]{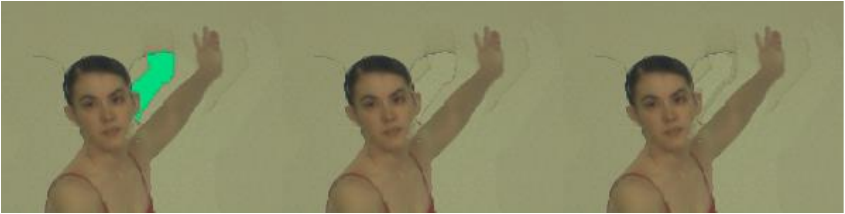}}   }
   \vspace{-0.25cm}
   \addtocounter{subfigure}{-1}
   \centerline{\subfigure{\includegraphics[width=1\hsize]{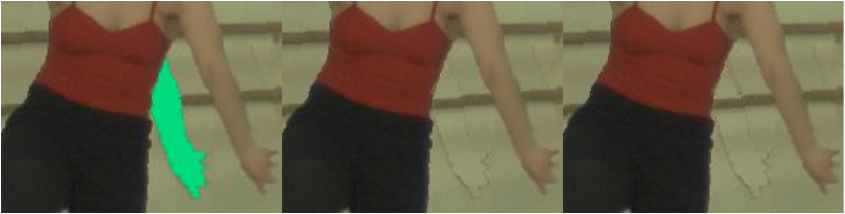}}   }
   \vspace{-0.25cm}
   \addtocounter{subfigure}{-1}
   \centerline{\subfigure{\includegraphics[width=1\hsize]{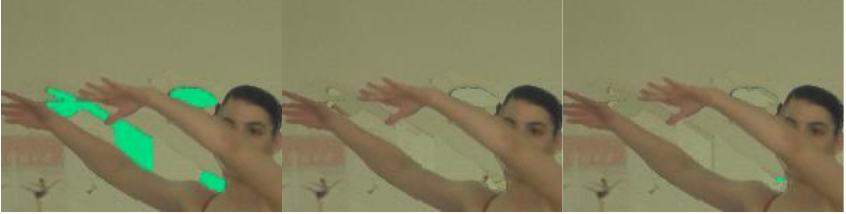}}   }
   \vspace{-0.25cm}
   \centerline{\subfigure[\textit{Ballet}]{\includegraphics[width=1\hsize]{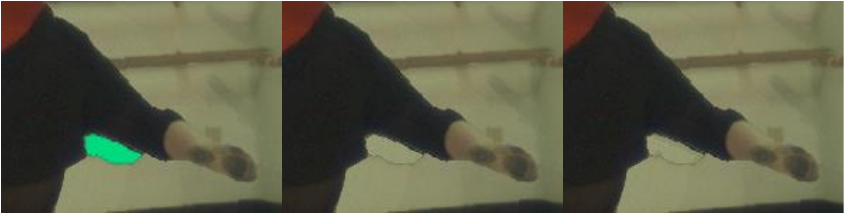}}   }
   \vspace{-0.1cm}

  \addtocounter{subfigure}{-1}
   \centerline{\subfigure{\includegraphics[width=1\hsize]{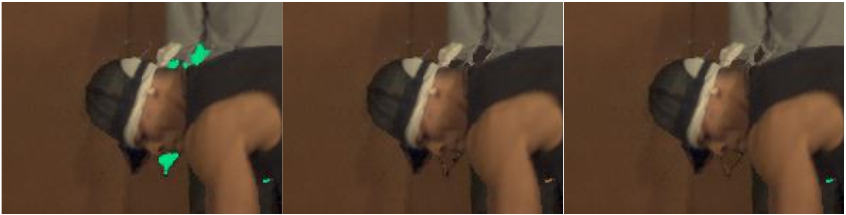}}   }
   \vspace{-0.25cm}
   \centerline{\subfigure[\textit{Breakdancers}]{\includegraphics[width=1\hsize]{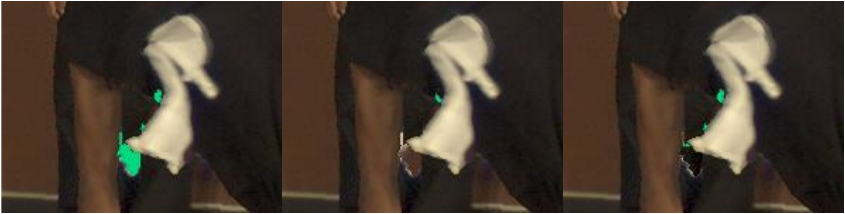}}   }
  \addtocounter{subfigure}{-1}
  %\vspace{-0.1cm}
   \centerline{\subfigure{\includegraphics[width=1\hsize]{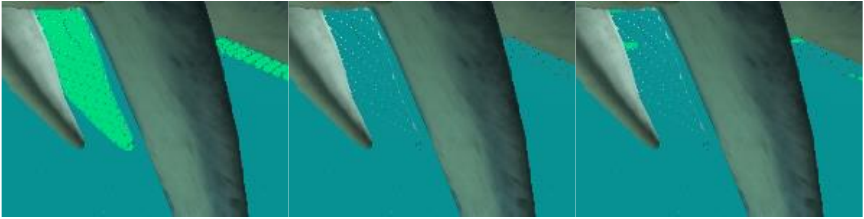}}   }
   \vspace{-0.25cm}
   \addtocounter{subfigure}{-1}
   \centerline{\subfigure{\includegraphics[width=1\hsize]{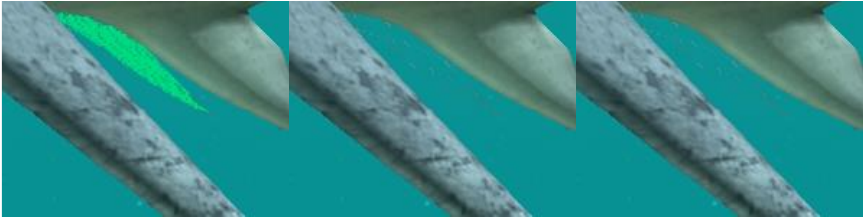}}   }
   \vspace{-0.25cm}
   \centerline{\subfigure[\textit{Shark}]{\includegraphics[width=1\hsize]{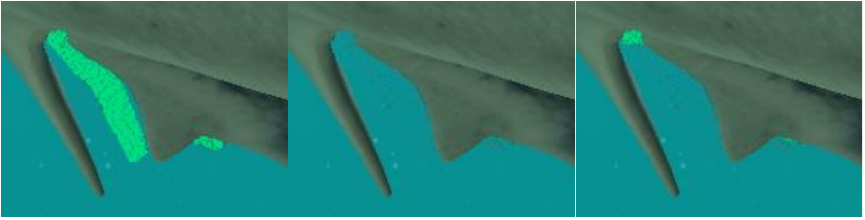}}   }

\end{figure}   
   
%\FloatBarrier

\newpage
\subsection{View Extrapolation}
The test sequences used for view extrapolation are listed in Table \ref{ve_ts} together with the specification of the target view to be synthesized, primary views and complementary views for reference. Table \ref{ve_size} shows the hole sizes per frame in the merged view by using PV, PV+CV and PV+SW\_CV, as well as the hole reduction percentages by PV+CV and PV+SW\_CV over PV, respectively. It can be seen that using multiple views for view extrapolation can achieve 26.5\% hole reduction in average, while the selective warping scheme can achieve 16.53\% hole reduction in average. The performance is not as good as that in view interpolation, with the reason that in view extrapolation most of the holes are generated in the case of single depth discontinuity (in view interpolation, these holes will be compensated by the other warped view). It can also be seen that the performance of view extrapolation using multiple views differs much among different sequences, indicating that the depth dynamic of the scene and the baseline length of each sequence are the key elements in determining the performance of view extrapolation. Table \ref{ve_psnr} shows quality comparison of the filled pixels by the proposed methods and image inpainting with respect to the original image. It can be seen that our proposed method achieves much better quality than the image inpainting scheme in terms of PSNR.

\begin{table}[t]
   \caption{\textsc{View extrapolation: test sequences and settings.}}\label{ve_ts}
   \centering
   \begin{tabular}{c c c c }%|p{4em}|p{3em}|p{3.5em}|p{3.5em}|p{3.5em}|p{3.5em}|
	\hline
	Sequence & Target View & Primary Views & Complementary Views\\
	\hline
	\textit{Ballet} & View 7	& View 5 & View 3\\
	\hline
	\textit{Breakdancers} & View 7	& View 5 & View 3\\
	\hline
	\textit{UndoDander} & View 9	& View 5 & View 1\\
	\hline
	\textit{Shark} & View 90	& View 75 & View 60\\
	\hline
	\textit{Bee} & View 100	& View 75	& View 50 \\
	\hline
	\end{tabular}
   
\end{table}
\begin{table}[t]
   \caption{\textsc{View extrapolation: hole size comparison by different approaches in terms of average number of pixels in holes (per frame).}}\label{ve_size}
   \centering
   \begin{tabular}{c | c c c | c c}%|p{4em}|p{3em}|p{3.5em}|p{3.5em}|p{3.5em}|p{3.5em}|
	\hline
	 \multirow{3}{*}{Sequence} & \multicolumn{3}{c|}{Hole Size}  & \multicolumn{2}{c}{Hole Reduction (\%)} \\ 
	 \cline{2-6}
	 & \multirow{2}{*}{PV} & PV+ & PV+ & PV+ & PV+  \\
	  & & CV & SW\_CV &  CV & SW\_CV \\ 
	\hline
	\textit{Ballet} & 99781	& 30208	& 67305	& 69.73	& 32.55\\
	\hline
	\textit{Breakdancers}& 24372	& 15720	& 17402	& 35.50	& 28.60 \\
	\hline
	\textit{UndoDancer}& 30058	& 28769	& 29240	& 4.29	& 2.72 \\
	\hline
	\textit{Shark}& 60304	& 54065	& 54586	& 10.35	& 9.48 \\
	\hline
	\textit{Bee} & 304112	& 265789	& 275889 & 12.6	& 9.28 \\
	\hline
	\end{tabular}
   
\end{table}
\begin{table}[t]
   \caption{\textsc{View extrapolation: PSNR comparison for the reduced hole pixels by PV+CV and PV+SW\_CV.}}\label{ve_psnr}
   \centering
   \begin{tabular}{c | @{\hskip 0.18in}  c @{\hskip 0.25in} c @{\hskip 0.18in}  | @{\hskip 0.18in}  c c }%|p{4em}|p{3em}|p{3.5em}|p{3.5em}|p{3.5em}|p{3.5em}|
	\hline
	 \multirow{3}{*}{Sequence} & \multicolumn{4}{c}{PSNR of Filled Pixels by Two Testing Approaches (dB)} \\ 
	 \cline{2-5}
	 & PV+ & PV+ & PV+ & PV+  \\
	  & Inpainting & CV & Inpainting & SW\_CV \\ 
	\hline
	\textit{Ballet} & 17.88	& 23.71	& 19.02	& 25.45\\
	\hline
	\textit{Breakdancers} & 19.25	& 21.52	& 19.23	& 21.52 \\
	\hline
	\textit{UndoDancer}& 11.5	& 17.88	& 12.85	& 21.83 \\
	\hline
	\textit{Shark} & 21.91	& 24.41	& 22.18	& 25.4 \\
	\hline
	\textit{Bee} & 15.47	& 17.02	& 15.38	& 18.16 \\
	\hline
	\end{tabular}
   
%\end{table}
\bigskip 
%\begin{table}[t]
   \caption{\textsc{View extrapolation: complexity comparison by different approaches in terms of average rendering time in sec (per frame).}}\label{ve_complexity}
   \centering
   \begin{tabular}{c c c c }%|p{4em}|p{3em}|p{3.5em}|p{3.5em}|p{3.5em}|p{3.5em}|
	\hline
	Sequence & PV & PV+CV & 2PV+SW\_CV\\
	\hline
	\textit{Ballet} & 0.178	& 0.343	& 0.231\\
	\hline
	\textit{Breakdancers}& 0.192	& 0.369	& 0.229\\
	\hline
	\textit{UndoDancer}& 0.5	& 0.976	& 0.584\\
	\hline
	\textit{Shark}& 0.495	& 0.956	& 0.582\\
	\hline
	\textit{Bee} & 0.468	& 0.889	& 0.64 \\
	\hline
	\end{tabular}
   
%\end{table}
\bigskip 
%\begin{table}[t]   
   \caption{\textsc{View extrapolation: hole size comparison by different approaches (using an extra complementary view) in terms of average number of pixels in holes (per frame).}}\label{ve_2_size}
   \centering
   \begin{tabular}{c | c c c | c c}%|p{4em}|p{3em}|p{3.5em}|p{3.5em}|p{3.5em}|p{3.5em}|
	\hline
	 \multirow{3}{*}{Sequence} & \multicolumn{3}{c|}{Hole Size}  & \multicolumn{2}{c}{Hole Reduction (\%)} \\ 
	 \cline{2-6}
	 & \multirow{2}{*}{PV} & PV+ & PV+ & PV+ & PV+  \\
	  & & 2CV & SW\_2CV &  2CV & SW\_2CV \\ 
	\hline
	\textit{Ballet} & 99781	& 10745	& 57988	& 89.23	& 41.88\\
	\hline
	\textit{Breakdancers}& 24372	& 10435	& 14301	& 57.18	& 41.32 \\
	\hline
	\textit{Shark}& 60304	& 19262	& 20310	& 20.15	& 15.8 \\
	\hline
	\textit{Bee} & 304112	& 246949	& 266387	& 18.8	& 12.4 \\
	\hline
	\end{tabular}   
\end{table}

\begin{figure}[htb]
  \centering
\includegraphics[width=0.83\hsize]{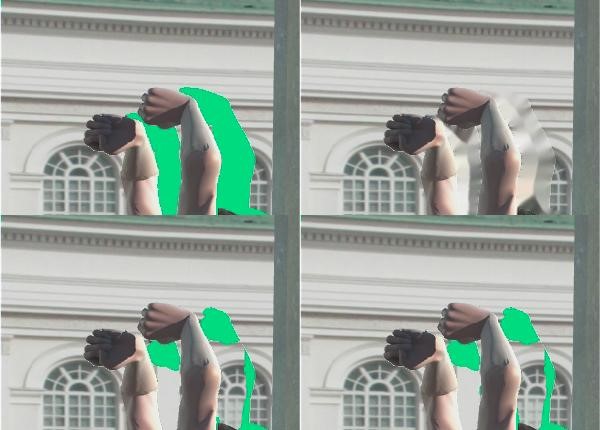}   
   \caption{Comparison of the synthesized views (snapshots of \textit{UndoDancer}) using different approaches. From left to right and top to down, the pictures are the synthesized views using PV, PV+image inpainting, PV+CV, and PV+SW\_CV, respectively, with holes in green.}
   \label{fig_prop_inpaint_snapshots}
\end{figure}

Some snapshots of different synthesized images by PV, PV+CV and PV+SW\_CV are shown in Fig. \ref{fig_ve_snapshots}. The pictures from left to right in each row denote the virtual views synthesized with PV, PV+CV and PV+SW\_CV, respectively, where holes are indicated in green. It can be seen that holes in the synthesized view by using the complementary view can be greatly reduced and in some cases are even completely eliminated. A comparison of the synthesized views (snapshots of \textit{UndoDancer}) using image inpainting and the proposed methods is shown in Fig. \ref{fig_prop_inpaint_snapshots}. As shown in the figure, for the hole between the two hands, there is not much relevant information around to help fill up the hole and consequently the filled pixels by image inpainting based on the irrelevant neighboring information deviate from the original ones. On the other hand, our proposed method reduces holes using pixels of other camera captured views based on the DIBR warping in the same way as other non-hole pixels, thus showing a better quality. Table \ref{ve_complexity} shows the complexity comparisons in terms of rendering time per frame in average in the same way as in Table \ref{vi_complexity}. It can be seen that the time increase percentage is very similar as that of view interpolation. 

Experiments on using even farther complementary views to assist the view extrapolation are also conducted where the results are tabulated in Table \ref{ve_2_size}. As can be seen from Table \ref{ve_2_size}, \textit{Ballet}, \textit{Breakdancers}, \textit{Shark} and \textit{Bee} are used for testing and the extra complementary views for the sequences are View 1, View 1, View 30 and View 25, respectively. In comparison against Table \ref{ve_size}, it can be seen that using one additional view in view extrapolation, about 10\% in average of hole pixels are reduced additionally.
%\FloatBarrier

%\newpage
\section{Conclusion}
\label{conclusion}
This paper has examined the hole generation and reduction in view interpolation and extrapolation using multiple reference views. It is shown that hole sizes in a synthesized view based on two primary views in view interpolation and one primary view in view extrapolation can be reduced by warping of the complementary views under certain mild conditions. Specifically, the conditions for hole reduction and the lengths of the reduced holes in both cases have been obtained. Accordingly, we have proposed a new view synthesis framework to synthesize virtual views using all the available reference views, which may be significantly useful for the 3DV and ongoing FTV project. In this framework complementary views have been warped to help reduce holes for high quality synthesized views. Furthermore, to lower the complexity of fully warping the complementary views, a selective warping scheme has been developed by locating a small portion of relevant pixels in the complementary views for hole filling. Experimental results have demonstrated that the proposed framework can effectively and efficiently reduce the hole sizes in both view interpolation and view extrapolation.

\newcommand\mhf{0.85}
\begin{figure}[!htbp]
  \centering
  \addtocounter{subfigure}{-1}
   \centerline{\subfigure{\includegraphics[width=\mhf\hsize]{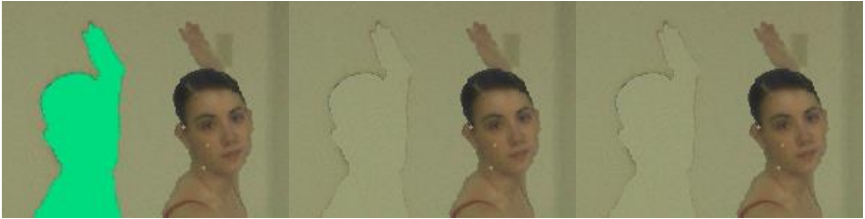}}   }
   \vspace{-0.3cm}
   \addtocounter{subfigure}{-1}
   \centerline{\subfigure{\includegraphics[width=\mhf\hsize]{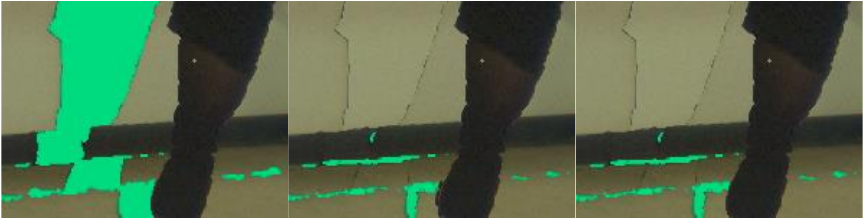}}   }
   \vspace{-0.3cm}
%   \addtocounter{subfigure}{-1}
%   \centerline{\subfigure{\includegraphics[width=\mhf\hsize]{Fig_10_a_3.pdf}}   }
%   \vspace{-0.3cm}
   \centerline{\subfigure[\textit{Ballet}]{\includegraphics[width=\mhf\hsize]{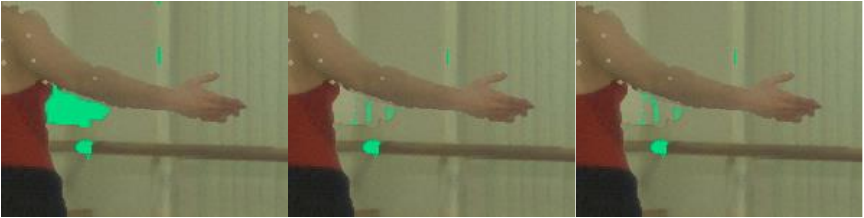}}   }
   \vspace{-0.3cm}
   
  %\centering
  \addtocounter{subfigure}{-1}
   \centerline{\subfigure{\includegraphics[width=\mhf\hsize]{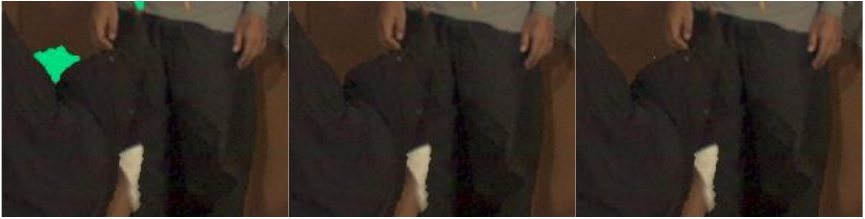}}   }
   \vspace{-0.3cm}
   
   \addtocounter{subfigure}{-1}
   \centerline{\subfigure{\includegraphics[width=\mhf\hsize]{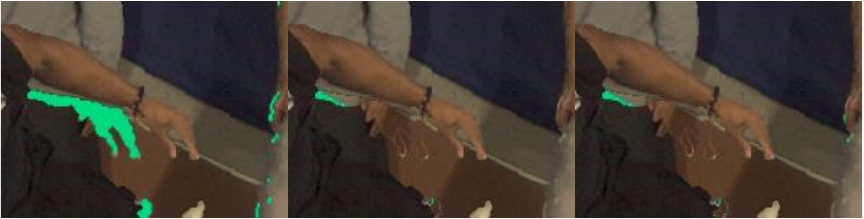}}   }
   %\vspace{-0.3cm}

\end{figure}
\begin{figure}[!htbp]
   \centerline{\subfigure[\textit{Breakdancers}]{\includegraphics[width=\mhf\hsize]{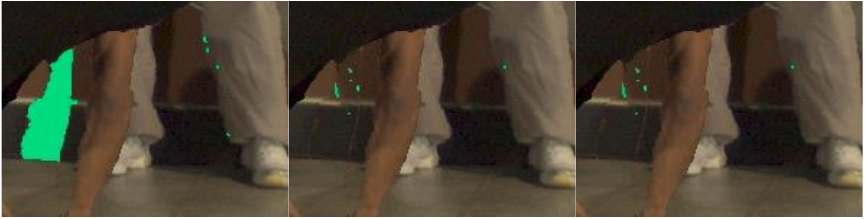}}   }
   \vspace{-0.3cm}
%\end{figure}
%\begin{figure}[!htbp]
%  \centering
  \addtocounter{subfigure}{-1}
   \centerline{\subfigure{\includegraphics[width=\mhf\hsize]{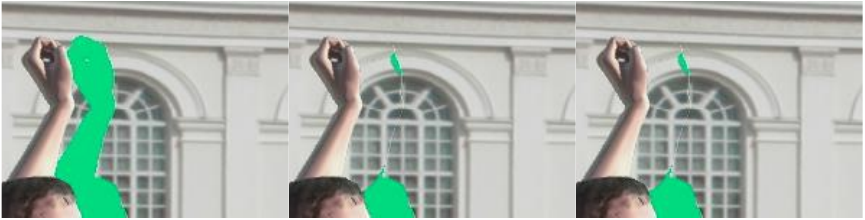}}   }
   \vspace{-0.3cm}
   \addtocounter{subfigure}{-1}
   \centerline{\subfigure{\includegraphics[width=\mhf\hsize]{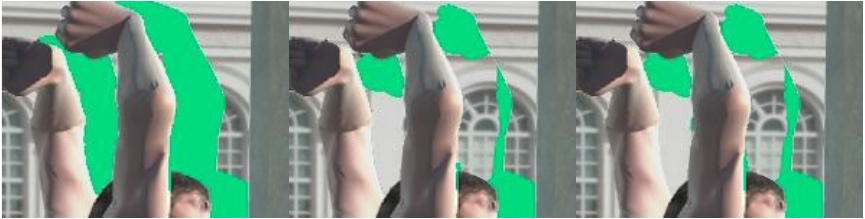}}   }
   \vspace{-0.3cm}
   \centerline{\subfigure[\textit{UndoDancer}]{\includegraphics[width=\mhf\hsize]{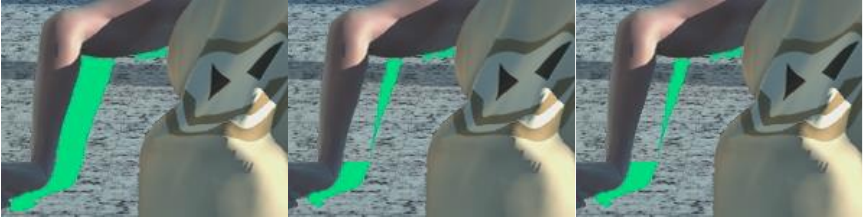}} }
  \centering
  \vspace{-0.3cm}
  \addtocounter{subfigure}{-1}
   \centerline{\subfigure{\includegraphics[width=\mhf\hsize]{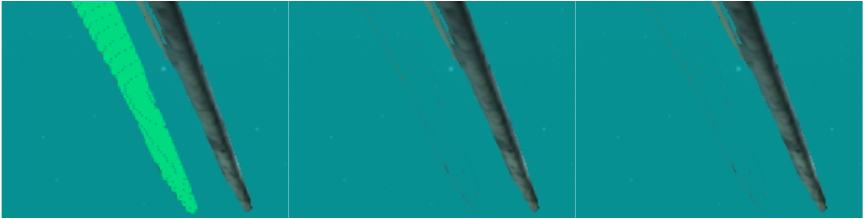}}   }
   \vspace{-0.3cm}
   \addtocounter{subfigure}{-1}
   \centerline{\subfigure{\includegraphics[width=\mhf\hsize]{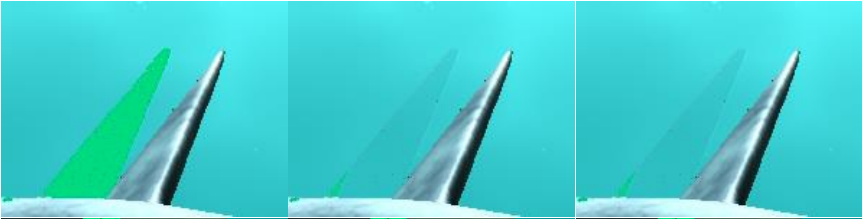}}   }
   \vspace{-0.3cm}
   \centerline{\subfigure[\textit{Shark}]{\includegraphics[width=\mhf\hsize]{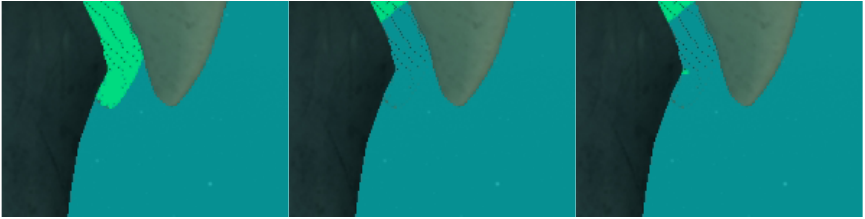}}  }
   \vspace{-0.3cm}
  \addtocounter{subfigure}{-1}
   \centerline{\subfigure{\includegraphics[width=\mhf\hsize]{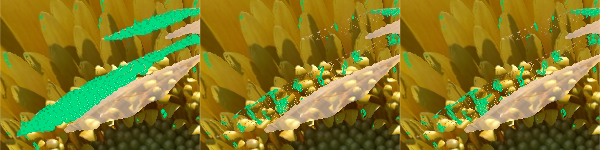}}   }
   \vspace{-0.3cm}
   \addtocounter{subfigure}{-1}
   \centerline{\subfigure{\includegraphics[width=\mhf\hsize]{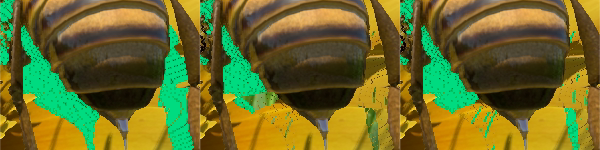}}   }
   \vspace{-0.3cm}
   \centerline{\subfigure[\textit{Bee}]{\includegraphics[width=\mhf\hsize]{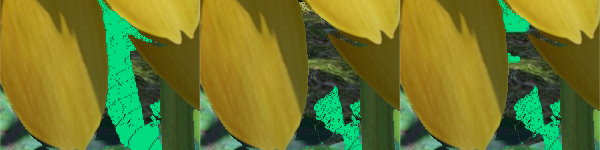}}   }  
   \vspace{-0.3cm}
   \caption{Snapshots of the synthesized views. The pictures from left to right are the synthesized views using PV, PV+CV and PV+SW\_CV, respectively, with holes in green.}
   \label{fig_ve_snapshots}
\end{figure}
%\FloatBarrier

\begin{IEEEbiography}[{\includegraphics[width=1in,height=1.25in,clip,keepaspectratio]{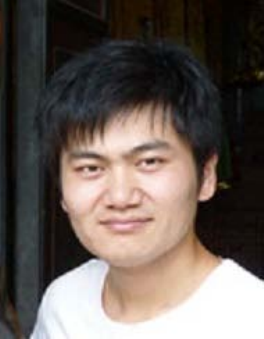}}]{Shuai Li}
is pursuing the Ph.D. degree in computer science with the University of Wollongong, Australia. He received the B.Eng. degree from Yantai University, China, in 2011, and the M.Eng. degree from Tianjin University, China, in 2014. He was with the University of Electronic Science and Technology of China as a Research Assistant from 2014 to 2015. His research interests include image/video coding, 3D video processing, and computer vision. He was a co-recipient of two best paper awards at the IEEE BMSB 2014 and IIH-MSP 2013, respectively.
\end{IEEEbiography}

\begin{IEEEbiography}[{\includegraphics[width=1in,height=1.25in,clip,keepaspectratio]{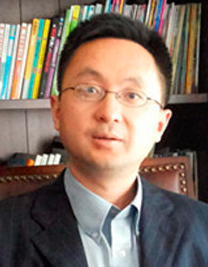}}]{Ce Zhu}
(M'03--SM'04--F'17) received the B.S. degree from Sichuan University, Chengdu, China, in 1989, and the M.Eng and Ph.D. degrees from Southeast University, Nanjing, China, in 1992 and 1994, respectively, all in electronic and information engineering. He held a post-doctoral research position with the Chinese University of Hong Kong in 1995, the City University of Hong Kong, and the University of Melbourne, Australia, from 1996 to 1998. He was with Nanyang Technological University, Singapore, for 14 years from 1998 to 2012, where he was a Research Fellow, a Program Manager, an Assistant Professor, and then promoted to an Associate Professor in 2005. He has been with University of Electronic Science and Technology of China, Chengdu, China, as a professor since 2012.

His research interests include video coding and communications, video analysis and processing, 3D video, visual perception and applications. He has served on the editorial boards of a few journals, including as an Associate Editor of \textsc{IEEE Transactions on Circuits and Systems for Video Technology}, \textsc{IEEE Transactions on Image Processing}, \textsc{IEEE Transactions on Broadcasting}, \textsc{IEEE Signal Processing Letters}, an Editor of \textsc{IEEE Communications Survey and Tutorials}, and an Area Editor of \textsc{Signal Processing: Image Communication}.

He is a co-recipient of two best paper awards and two best student paper awards at the IIH-MSP 2013, the IEEE BMSB 2014, the MobiMedia 2011, and the IEEE ICME 2017, respectively.
\end{IEEEbiography}

\begin{IEEEbiography}[{\includegraphics[width=1in,height=1.25in,clip,keepaspectratio]{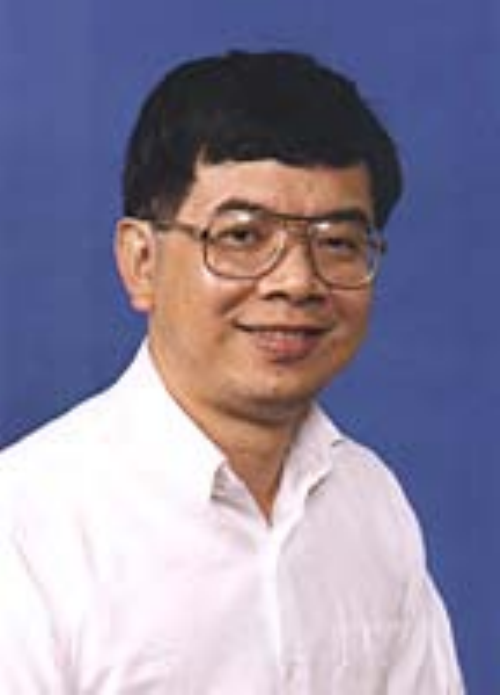}}]{Ming-Ting Sun} (S'79--M'81--SM'89--F'96) received the B.S. degree from National Taiwan University, Taipei, Taiwan, and the Ph.D. degree from the University of California at Los Angeles, Los Angeles, all in electrical engineering. He joined the University of Washington, Seattle, in 1996, where he is currently a Professor. He was the Director of the Video Signal Processing Research Group, Bellcore, NJ.

He is a Chaired or Visiting Professor with Tsinghua University, Tokyo University, National Taiwan University, National Cheng Kung University, National Chung Cheng University, National Sun Yat-sen University, the Hong Kong University of Science and Technology, and National Tsing Hua University. He holds 13 patents and has published over 200 technical papers, including 17 book chapters in video and multimedia technologies. He is the co-editor of the book \textit{Compressed Video Over Networks}. His current research interests include video and multimedia signal processing, video coding, computer vision, and machine learning.

Dr. Sun was an Editor-in-Chief of \textsc{Journal of Visual Communication and Image Representation} from 2012 to 2016. He was the Guest Editor of 11 special issues for various journals, and has given keynotes for several international conferences. He was a member of many prestigious award committees and was a Technical Program Co-Chair of several conferences including International Conference on Multimedia and Expo 2010. He served as the Editor-in-Chief of the \textsc{IEEE Transactions on Multimedia} and a Distinguished Lecturer of the Circuits and Systems Society from 2000 to 2001. He received the IEEE CASS Golden Jubilee Medal in 2000. He was the Editor-in-Chief of the \textsc{IEEE Transactions on Circuits and Systems for Video Technology} from 1995 to 1997. He received the \textsc{IEEE Transactions on Circuits and Systems for Video Technology} Best Paper Award in 1993. From 1988 to 1991, he was the Chairman of the IEEE CAS Standards Committee, and established the IEEE Inverse Discrete Cosine Transform Standard. He received the Award of Excellence from Bellcore for his work on the digital subscriber line in 1987.
\end{IEEEbiography}

%\end{IEEEbiography}

%% insert where needed to balance the two columns on the last page with
%%\newpage


%\vfill


\end{document}